# Rationality, Cooperation and Conversational Implicature

**Mark Lee**
Department of Computer Science
University of Sheffield
Regent Court, 211 Portobello Street
Sheffield S1 4DP, UK
*M.Lee@dcs.shef.ac.uk*

**Abstract**

Conversational implicatures are usually described as being licensed by the disobeying or flouting of a *Principle of Cooperation*. However, the specification of this principle has proved computationally elusive. In this paper we suggest that a more useful concept is rationality. Such a concept can be specified explicitly in planing terms and we argue that speakers perform utterances as part of the optimal plan for their particular communicative goals. Such an assumption can be used by the hearer to infer conversational implicatures implicit in the speaker's utterance.

## 1. Introduction

Previous work in pragmatics has relied upon a *Principle of Cooperation* [Grice, 1975] to explain how speakers can communicate more than what is explicitly said. Despite, the dominance of such work in the philosophy of language, the principle remains too computationally underspecified, and therefore, unused by dialogue systems in artificial intelligence (AI). Furthermore, the actual process of inferring an implicature due to the speaker disobeying the Principle is unclear.

In this paper, we present an alternative theory of conversational implicature based on the concept of rationality. Rather than assuming that speakers are cooperative, the theory argues that speakers achieve their conversational goals as *rational agents*. This theory is implemented in ViewGen, a computer program which models the propositional attitudes of agents engaged in dialogue. ViewGen is implemented in Quintus prolog and currently we are investigating its use in a medical counselling domain.

Our theory is summarised as follows: previous accounts of dialogue understanding have assumed that dialogues are cooperative so that the participants are truthful, informative but not verbose, relevant and clear. While speakers do in general follow these maxims, the underlying reason for this is that they are rational agents who, when possible, achieve their goals using *optimal plans*. Given that a speaker performs an utterance, if the hearer recognises a non-optimal plan, then there must be a conversational implicature which the speaker intends to communicate. Understanding this implicature involves understanding why this particular plan was chosen and acted upon as opposed to the apparently optimal plan for the explicit communicative goal.

## 2. Cooperation and pragmatic understanding

According to Grice, conversational implicatures arise due to the set of assumptions that exist in language use. More specifically, Grice identifies a Principle of Cooperation which instructs language users to:

> "make [their] conversational contribution such as is required, at the stage at which it occurs, by the accepted purpose or direction of the talk exchange in which you are engaged." [Grice, 1975]:45

In order to flesh out this principle, Grice suggests four general maxims which if observed will fulfil the cooperative principle: the maxims of Quality (truthfulness), Quantity (no more and no less than required), Relation (relevance) and Manner (presentation). Grice's position is that after the hearer has recognised the apparent flouting of a maxim, he or she draws an inference that the speaker is communicating an additional implicature which explains why the maxim was disobeyed.

However, the Principle of Cooperation has not proven to be useful in building dialogue understanding systems. The most apparent problem is that the maxims are not specified sufficiently for use by computers. However, there are reasons to believe the explicit representation of the position is not required. For example, Dale and Reiter [Dale and Reiter, 1995] argue that any reasonable natural language generation (NLG) system will obey the maxims anyway without an explicit representation of the maxims. Simply, any well designed NLG system will only produce truthful contributions satisfying all and only the set of communicative goals available which are relevant to the system in as clear a manner as possible.

However, another purpose for the Principle of Cooperation is to provide an account for when an utterance flouts a maxim to communicate an implicature. However, it is not clear how the recognition of an utterance breaking the Principle leads to the inference of the correct implicature. In fact, there has been very little computational work on the processing of implicatures from the flouting of Grice's maxims. In the next section, we will develop an alternative approach based on *rationality* and then in the remainder of the paper show how it can be used to understand conversational implicatures.

## 3. Rationality

Rational agents ideally achieve their goals by the use of optimal plans. Research in plan generation has specified a number of heuristics for producing such plans. Specifically, good plans have the following criteria:

**Correctness:**
All actions in the plan should rely on correct propositions at the time of their execution. In terms of agent modelling, the criterion of correctness requires that agents prefer plans which are grounded on propositions which are believed by the agent to be true.

**Relevance**
The plan has a whole achieves the complete set of goals required by the planning agent. The criterion of relevance dictates that agents plan to achieve the maximum set of goals they can with the plan.

**Efficiency**
The plan achieves the stated goals incurring the minimum cost in terms of either time or effort or resources used. The criterion of efficiency dictates that agents prefer the cheapest plan available. Cost can refer to time, effort or resources. For the purposes of this paper, a simple measure of effort based on the number of planning steps is sufficient.

Clearly there is a tension between the three criteria. Typically, correct plans require more specification than abstract plans. This additional specification increases the planning cost expressed in both time and effort and therefore conflicts with the efficiency criterion. However, in this paper, we will concentrate on the tension between the criteria of relevance and efficiency. The inference of implicatures based on the correctness criterion is discussed further in [Lee and Wilks, 1997].

The criterion of relevance suggests that a good plan should achieve as large a number of goals for the planner as possible. Stated simply, the more goals that are achieved, the more "relevant" the plan, and therefore, the more relevant the action performed to the agent. However, the criterion of efficiency suggests that a good plan should be inexpensive and, therefore, for a given set of goals, the shorter plan should be preferred over the longer plan with all other things being equal.

In plan generation, given a fixed set of goals, a simple heuristic is to generate the shortest plan for the full set of goals to ensure that both criteria are satisfied. However, during plan recognition, this is more difficult since plan recognition involves inferring the actual set of goals the speaker is trying to achieve. The size of this set is usually unknown and therefore, it is not clear when either of the above criteria is satisfied by the recognised plan.

In the following sections, we will present an account where additional goals are ascribed to the speaker of an utterance (thus increasing the relevance of the utterance) based on the assumed rationality of the speaker and therefore, his or her intent to maximise the efficiency of the plan behind the utterance.

This theory is part of a larger theory of belief modelling and ascription implemented within the ViewGen system. ViewGen is briefly described in the next section.

## 4. ViewGen

ViewGen [Wilks et al., 1991] is a dialogue understanding system which reasons about the attitudes of other agents in nested belief structures. It does this by *ascription* - i.e. assuming that attitudes held in one attitude box can be ascribed to others. There are two main methods of ascription - default ascription and stereotypical ascription. Default ascription applies to common attitudes which ViewGen assumes that any agent will hold and also ascribe to any other agent unless there is contrary evidence. Stereotypical ascription applies to stereotypical attitudes which ViewGen assumes apply to instances of a particular stereotype e.g.: ViewGen ascribes expert medical knowledge to doctors. Stereotypes can also apply to types of dialogue. For example, knowledge elicitation goals can be ascribed to agents involved in information seeking dialogues. Where as default ascriptions can be applied to any agent, stereotypical ascription requires the precondition of some trigger for the stereotype to be applied. For example, the use of a particular speech act stereotypically assumes the holding of certain attitudes. The use of stereotypes allows ViewGen to order candidate goals in terms of their likelihood during plan recognition.

ViewGen is capable of recognising and using a set of speech acts [Lee and Wilks, 1996]. The acts are specified as plan operators which are used to plan what is said and recognise dialogue plans from the user. For example, informing is specified as:

**Inform(Speaker,Hearer,Proposition)**
**Preconditions: goal(Speaker,bel(Hearer,Proposition))**
**bel(Speaker,Proposition)**

where the predicates goal and bel refer to goals and beliefs. Rather than specify the effects of an act, there are separate ascription rules for the speaker and hearer of any act:

<u>Update on the Speaker's belief set</u>
  For every condition C in a dialogue act performed:
    *ascribe(Speaker, Hearer, bel(C))*
<u>Update on the Hearer's belief set</u>
  For every condition C in a dialogue act performed:
    *ascribe(Hearer, Speaker, C)*

This account differs from standard accounts of computational speech acts (e.g. [Allen, 1983]) in that the underlying operation of ascription takes over implicitly many of the conditions and effects that are specified explicitly in other approaches. For example, the Inform act given above creates the following belief ascriptions on the part of the hearer:

**bel(Hearer, goal(Speaker,bel(Hearer,Proposition)))**
**bel(Hearer, bel(Speaker,Proposition))**

Therefore, rather than define speech acts by their effect, they are defined solely in terms of their conventional preconditions and their success or failure depends on the communication of these belief based preconditions. This allows acts to be planned by their minimal, context free effects. Further effects due to the context, e.g. whether the Inform act actually updates the hearer's personal belief of the communicated proposition, are separate from the act and modelled as further inferences or ascriptions in the dialogue plan. For example, there is an Accept_belief plan operator which allows the hearer to accept a speaker's belief if 1) there is no contrary evidence; 2) the speaker is a reliable source of information (i.e. an expert in the domain in question etc.). This allows a clear representation of how the conventional meaning of an utterance interacts with the context to produce a non-conventional effect such as in the case of an indirect speech act. Further details of how ViewGen represents and processes speech acts are given in [Lee and Wilks, 1996].

## 5. Licensing implicatures from plan inefficiencies

As noted in Section 3, the criteria of relevance and efficiency typically conflict. A maximally relevant plan is one which attempts to find a solution to all of the speaker's current goals yet typically such a plan will be longer than a shorter plan which concerns only a subset of the goals of the speaker. During plan recognition initially, the speaker is assumed to be trying to achieve one task related goal. Plan recognition proceeds as follows: given an utterance, its minimal meaning is derived from its surface form. An utterance's minimal meaning is the set of communicated attitudes associated with the speech act of the utterance. The planner then attempts to generate a plan which connects the utterance with one ascribable goal. In the current implementation of ViewGen, we use a partial order clausal link planner [McAllester and Rosenblatt, 1991] to generate such a plan.

The initial plan can be evaluated by re-planing a solution for the ascribed goal, bypassing the speaker's utterance. If the planner can find a plan which is shorter in length than the speaker's ascribed plan then the speaker's plan can be seen to be inefficient. In such a case, assuming the speaker is rational, there are two possibilities: either the speaker is attempting an additional *conjunctive goal* or bypassing an *avoidance goal.*

Work in the recognition of speech acts has generally assumed that only a single task related goal is intended by the speaker. However, in the present theory, though any utterance is associated with a single illocutionary speech act. However, its effects and its role in the larger dialogue plan might result in context specific effects and play a role in several separate communicative goals. For example, the answer below has the duel goals of offering an answer and a warning:

**Question: Can I swim in the sea today?**
**Answer: The waves are too strong and you'll drown.**

Such goals can be termed *conjunctive goals* where the speaker is attempting to satisfy additional communicative goals implicitly. In addition to the speaker attempting to satisfy positive goals such as informing the user of the context behind a request or requesting additional information, speakers often attempt to avoid certain topics. For example, consider in the context of cakes being burnt:

**Question: Have you checked the cakes in the oven?**
**Answer: I was watching the water boiling in the pan.**

Where the reply clearly is an attempt to avoid giving a direct answer to the question. Such goals can be termed *avoidance goals* where the speaker is attempting to avoid communicating a given topic or proposition.

## 6. Goal ascription after plan recognition

In this section, we provide a more formal treatment of conjunctive and avoidance goals. Consider if the speaker utters an utterance U. The hearer can recognise U as a certain illocutionary speech act by its surface features. From this speech act, a plan $P_r$ consisting of the actions $\{U, A_1...A_n\}$ can be recognised which achieves a single ascribable goal $G_1$. However, there exists a plan $P_o$ which also achieves $G_1$ such that

$$\text{cost}(P_o) < \text{cost}(P_r)$$

### Conjunction

Clearly, if $P_r$ is the plan chosen by the speaker then there must be some reason why it is preferred. As noted above, one possibility is that the speaker is attempting to achieve a conjunctive goal. Such a goal must be able to be planned from an exclusive state of the recognised plan and this updated plan should be the optimal solution i.e. the most efficient for achieving the conjunction of the goals.

If $G_2$ is the conjunctive goal then there must be a sub-plan or *completion* which if performed from some state exclusive to $P_r$ will achieve $G_2$. A completion is a sub plan i.e. a set of ordered actions. Such a completion requires that at least one precondition of the completion is enabled by an exclusive state in $P_r$. Such a completion is shown in Figure 1(a).

Once such a completion is found, the goal satisfied by the completion can be ascribed to the speaker as a conjunctive goal which the speaker intends to satisfy. In addition, the set of actions can be ascribed as an additional set of intentions which the speaker intends to perform. Such an ascription is possible if the following is true:

> **Ascription of a conjunctive goal**
> Given a goal $G_2$ which can be achieved by a completion of actions $\{A_1...A_n\}$ from some state S in $P_r$ where $P_r$ is the recognised plan which satisfies the goal $G_1$
> $$\text{default ascribe}(Hearer, Speaker, \text{goal}(G_2))$$
> and ascribe each intention in the completion
> $$\text{default ascribe}\{Hearer, Speaker, \text{int}(A_1...A_n))$$
> Condition of exclusiveness
> *The ascribed goal $G_2$ is achieved by the satisfaction of a state S which is exclusive to $P_r$*
> Condition of efficiency
> *The conjunction of the ascribed goal $G_2$ and the pre-existing ascribed goal $G_1$ is optimally satisfied by $P_r$ plus the completion from the exclusive state.*

### Avoidance

Another possibility is that the speaker is using the recognised plan because they are attempting to avoid a state which occurs in the apparently optimal plan. Such a state must lead to an avoidance goal. If AG is an avoidance goal then there must be a completion from an state exclusive to the optimal plan $P_o$ which leads to the satisfaction of the avoidance goal. Such a completion must contain only actions where the speaker plays no part as an actor. This is to ensure that once the state in Po is reached, the goal which the speaker wishes to avoid potentially could be satisfied without the speaker's intervention. In such a case, once the state is reached then the avoidance goal can be achieved regardless of the speaker's actions and so therefore, the speaker cannot allow the state to be achieved. Once such a completion is found, the negation of the avoided goal can be ascribed to the speaker as a goal. Such an ascription is possible if the following conditions are true:

> **Ascription of an avoidance goal**
> Given a goal AG which can be achieved by a completion of actions $\{A_1...A_n\}$ from some state in $P_o$ where $P_o$ is the optimal plan which satisfies the goal $G_1$
> $$\text{default ascribe}(Hearer, Speaker, \text{goal}(\text{not}(AG)))$$
> Condition of exclusiveness
> *The goal state AG is achieved by the satisfaction of a state S which is exclusive to $P_o$*
> Condition of causality
> *No action in the completion which satisfies AG from the state S features the Speaker as*

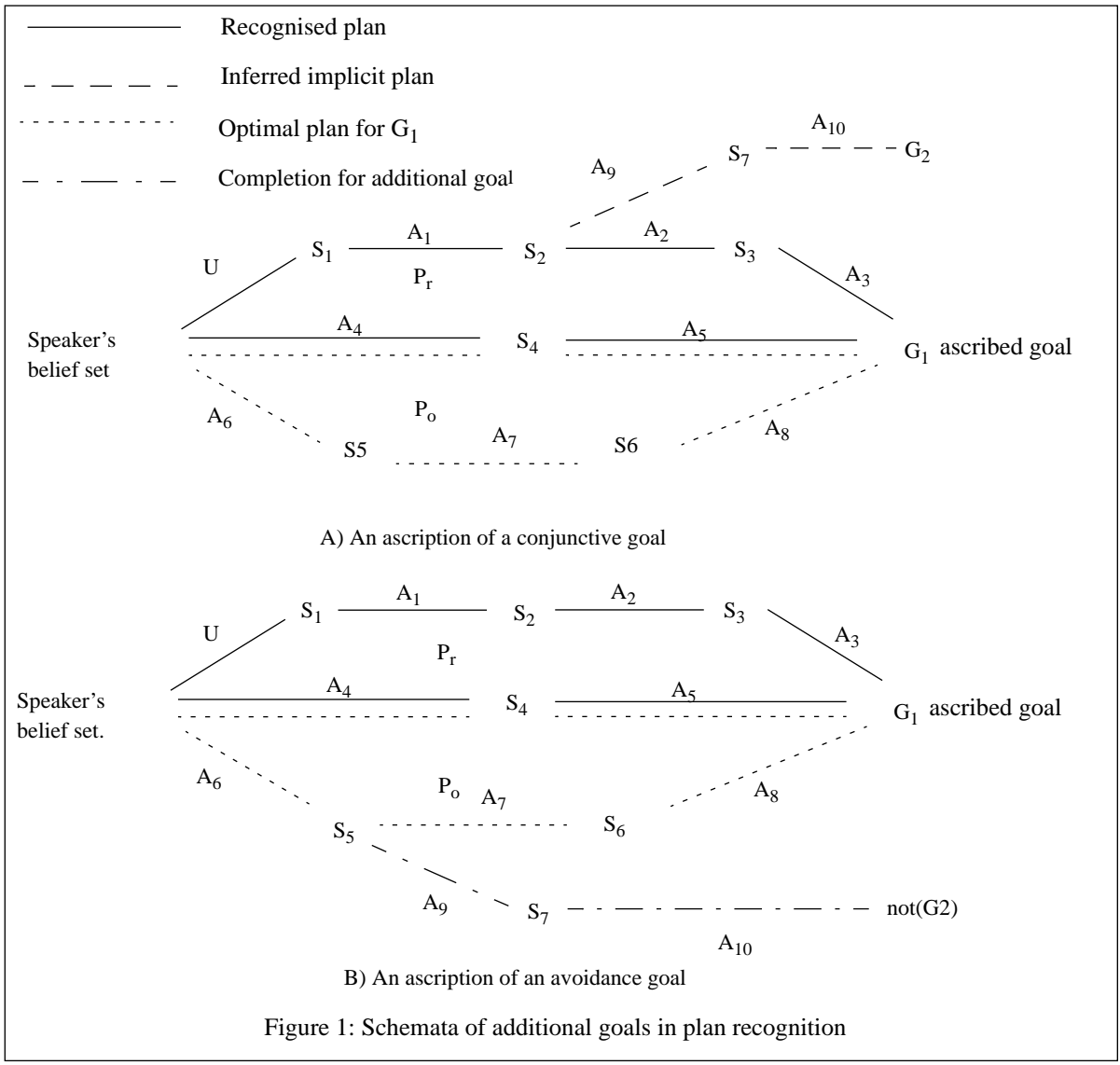

Figure 1: Schemata of additional goals in plan recognition

*an agent.*

Using the above two ascription rules, it is possible to recognise either case and ascribe the appropriate intentions and communicative goals, given a non-optimal plan. In the next section, we illustrate this by example.

## 7. Example

The above definitions can be used to recognise the speaker's intent to achieve more than one communicative goal with their utterance. Therefore, the hearer can ascribe additional goals to the speaker's plan to represent the contextual meaning of the performed speech act. Consider the following question and (indirect) answer exchange:

**System: Can I switch off the computer now?**
**Expert: You will damage the hard drive.**

In most contexts, such a reply by the Expert would be interpreted as an indirect negative answer. Green and Carberry [Gr+Ca92] note that in their corpus of question-answer pairs, such indirect replies constituted 10% of the answers. The System's question can be represented by the following speech act:

**question(System, Expert, permission(System, switch(System, computer off)))**

As discussed above, the performance of any speech act presupposes the holding of certain attitudes. For a question, the speaker should have a goal of wishing to believe the correct answer to the question and believe that the hearer knows this answer. In the context of the current utterance, the System should specifically believe the following attitudes[1]:

**goal(System, bel(System,permission(System, switch(System, computer off)) or**

---

1. In this paper, we represent yes-no questions by a true/false disjunction of the propositional content.

**bel(System, not(permission(System, switch(System, computer off)))))**

and

**bel(System, bel(Expert, permission(System, switch(System, computer off)) or bel(Expert, not(permission(System, switch(System, computer off))))**

That is, the System has the goal of believing whether or not the System can switch off the computer and believes that the Expert believes either that the System has or has not permission. The Expert replies to the question with a statement informing the System that such an action will damage the hard drive. This utterance can be represented by the following Inform act:

**inform(Expert, System, cause(switch(System, computer off), damage(hard drive)))**

As described above, the utterance and the ascribed belief set of the speaker (in this case, the Expert) can be used to infer a plan which achieves some stereotypical goal. Given the immediate discourse context of the System's performed question, there is an expectation that the Expert is providing either a yes or a no answer. Therefore, the two most likely candidate goal ascriptions are:

"yes_goal"

**goal(Expert, bel(System, permission(System, switch(System, computer off))))**

"no_goal"

**goal(Expert, bel(System, not(permission(System, switch(System, computer off)))))**

Therefore, if a plan can be constructed from the recognised Inform act to either goal then this plan and the specific goal will be ascribed to the Expert by the System. According to the rewrite rules for the effects of a speech act on a hearer, the Expert's Inform act has the following effects:

**(i) bel(System, bel(Expert, cause(switch(System, computer off), damage(hard drive))))**

**(ii) bel(System, goal(Expert, bel(System, cause(switch(System, computer off), damage(hard drive)))))**

A stereotypical goal held by computer experts is to ensure any states where computer equipment is damaged are avoided. If this stereotype is regarded as commonly accepted then the Expert can assume that the System is capable of stereotypically ascribing this goal to the Expert. This ascription can be represented by the operator:

**ascribe(System, goal(Expert, not(damage(hard drive))))**

Such an ascription achieves the following belief state:

**bel(System, goal(Expert, not(damage(hard drive))))**

Given that the Expert has the belief (i) that switching the computer off will damage the hard drive and the goal of avoiding this, the Expert must intend that the System does not perform such an action. Given, that these premises can be ascribed to the System, this belief can also be ascribed so that the System believes that the Expert intends the System not to switch the computer off:

**bel(System, int(Expert, not(switch(System, computer off))))**

Therefore, the Expert believes that the System should not have permission to switch the computer off. Since this chain of inference is being simulated in the Expert's belief environment of the System's beliefs, it results in the following belief update:

**bel(System, bel(Expert, not(permission(System, switch(System, computer off)))))**

Using an accept belief operator, the Expert can simulate the System accepting the above belief as a consequence of the utterance so that the System believes that it does not have permission to switch the computer off, i.e.:

**bel(System, not(permission(System, switch(System, computer off))))**

In doing so, the Expert would of achieved his or her goal of answering the System's question as a no_goal by the use of an Inform act to indirectly perform a No-answer speech act.

So far, the analysis provided is similar in approach, though not specifics, to other indirect speech act processing approaches to[1]. However, this plan can be further analysed to infer the additional communicative goals behind the Expert's intent.

A more efficient plan for achieving the above ascribed goal would be for the Expert simply to perform a No_answer speech act i.e.:

**no_answer(Expert, System, permission(System, switch(System, computer off)))**

resulting in the following effects:

**bel(System, bel(Expert, not(permission(System, switch(System, computer off)))))**

**bel(System, goal(Expert, bel(System, not(permission(System, switch(System, computer off))))))**

An accept-belief operator could then be used by the System to accept the act's first effect i.e. the belief that the System does not have permission to switch off the computer to achieve the state:

**bel(System, not(permission(System, switch(System, computer off))))**

which directly achieves the Expert's ascribed No-goal.

---

1. See [Cohen et al., 1990] for representative sample of such approaches.

Given the System's assumption of rationality on the part of the Expert, there must be some ulterior reason why the Expert choose the recognised plan over this "optimal" plan. If the criterion of relevance is assumed to have been fulfilled, then it is clear that the criterion of efficiency has been flouted since for the single ascribed goal, there exists a less expensive plan. Therefore, the Expert must be intending to achieve some additional goal.

As defined in Section 6, either a completion must exist from the optimal plan to some undesired state (i.e. an avoidance goal) or a completion from the preferred plan to an additional desired state (i.e. a conjunctive goal).

In the current example, a potential conjunctive goal exists. In ViewGen's domain model, there exists an environment of stereotypical computer expert goals. One such goal is the education of computer novices. Such a goal is defined to be achieved when the Expert causes the hearer to accept a correct belief associated with the Expert's area of knowledge. One such goal would clearly be the communication that switching the computer off damages the hard drive. For example:

"educate_goal"

**goal(Expert, bel(?, cause(switch(?, Computer off)), damage(hard drive))))**

Such a goal can be achieved from effect (i) of the Expert's Inform act, i.e. from the state:

**bel(System, bel(Expert, cause(switch(System, computer off)), damage(hard drive))))**

where the System believes the Expert believes the informed proposition. The educate goal can be achieved by the use of an accept_belief operator i.e.:

**accept_belief(System, Expert, cause(switch(?, Computer off)), damage(hard drive)))**

to achieve the following effect:

**bel(System, cause(switch(System, Computer off)), damage(hard drive))**

Such a plan completion can be added to the prior recognised plan. In addition, the intention for this belief to be accepted by the System can be ascribed as an intention to the Expert. The resulting, final plan of the Expert is shown in Figure 2.

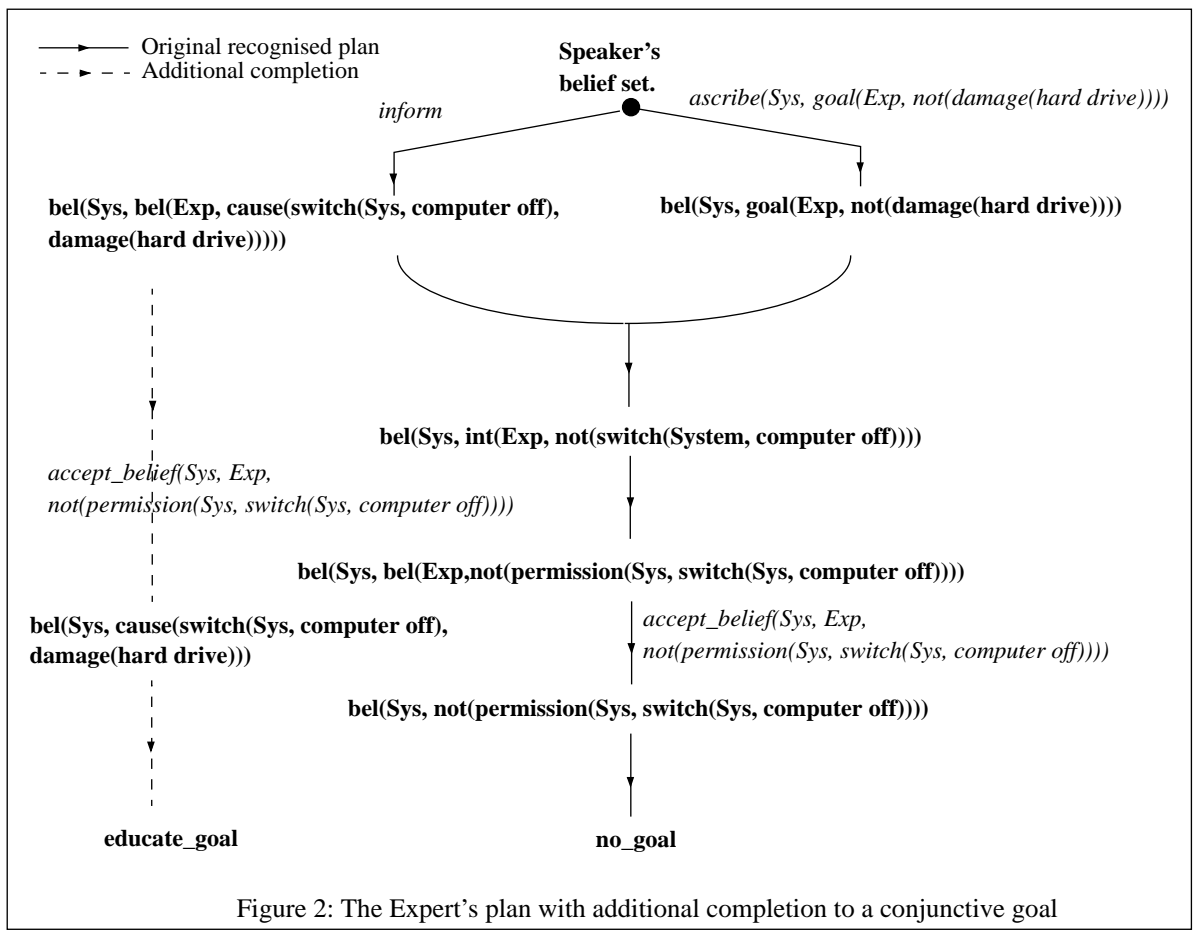

Figure 2: The Expert's plan with additional completion to a conjunctive goal

## 8. Related work

Pollack [Pollack, 1990] presents an account of plan recognition which deals with the possibility of mistaken beliefs on the part of the speaker by ascribing a set of beliefs associated with his or her plan plus a set of meta level beliefs explaining the role the first set of beliefs play in the speaker's plan. Pollack's idea of ascribing a set of additional beliefs based on the intentions contained in the recognised plan is similar to our idea of ascribing additional conversational goals to explain why a plan is apparently inefficient. However, her motivation is to show how mistaken beliefs in dialogue can be recognised and dealt with and is, therefore more concerned with the validity of the agent's beliefs rather than the overall rationality of their plans. In addition, she is not concerned with how conversational implicature can be understood.

Green and Carberry [Green and Carberry, 1992] provide an alternative analysis of indirect answers to yes-no questions. They represent turns in the modelled dialogue as plan operators encoding rhetorical structure theory relations. Such relations feature a nucleus which is the central communicative goal plus satellites which feature additional pieces of information which can be communicated. Green and Carberry argue that if a nucleus is missing but a satellite is given then the nucleus can be inferred as a conversational implicature. Green and Carberry's work provides a practical method of planning conversational implicature. However, we believe our work has the advantage that we derive implicature not from the higher level of rhetorical structures but a lower level of analysis based on beliefs, intentions and goals. This provides our approach with a more general method which can be applied to other classes of conversational implicature.

## 9. Discussion and Conclusions

In the sections above, we have shown a method of deriving conversational implicature based on the apparent irrationality of a speaker's plan behind his or her utterance. In our approach, conversational implicatures are not represented by the addition of semantic meaning but by the addition of conversational goals and intentions. Implicatures are recognised by the inefficiency of the speaker's dialogue plan. Such an inefficiency suggests that the utterance has a greater relevance to the speaker than just the initially recognised goal of the speaker. This greater relevance can be understood by additional goal ascription to the speaker's plan.

It is an open question whether our intuitions fully capture plan rationality or cover the full set of conversational implicatures captured by Grice's Principle. We hope that further work involving corpus analysis with provide further insight.